\pdfoutput=1
\documentclass[sigconf]{acmart}

\usepackage{graphicx}
\usepackage{tabularx}
\usepackage{multirow}
\usepackage{arydshln}
\usepackage{makecell}
\usepackage{pifont} 
\usepackage{balance} 
\usepackage{amsmath} 

\AtBeginDocument{%
  \providecommand\BibTeX{{%
    \normalfont B\kern-0.5em{\scshape i\kern-0.25em b}\kern-0.8em\TeX}}}

\setcopyright{acmlicensed}
\copyrightyear{2023}
\acmYear{2023}
\acmDOI{10.1145/3539618.3592043}

\acmConference[SIGIR '23]{the 46th International ACM SIGIR Conference on Research and Development in Information Retrieval}{July 23--27, 2023}{Taipei, Taiwan}
  
\acmBooktitle{Proceedings of the 46th International ACM SIGIR Conference on Research and Development in Information Retrieval (SIGIR '23), July 23--27, 2023, Taipei, Taiwan}
\acmPrice{15.00}
\acmISBN{978-1-4503-9408-6/23/07}

\begin{document}

\title[Short Research Paper]{Prompt Learning to Mitigate Catastrophic Forgetting in Cross-lingual Transfer for Open-domain Dialogue Generation}


\author{Lei Liu}
\orcid{0000-0003-3329-6795}
\affiliation{
    \department{Information Retrieval and Knowledge Management Research Lab}
    \department{Department of Electrical Engineering \& Computer Science}
    \institution{York University}
    \streetaddress{4700 Keele Street}
    \city{Toronto}
    \country{Canada}
    \postcode{M3J 1P3}}
\email{lliu@eecs.yorku.ca}

\author{Jimmy Xiangji Huang}
\orcid{0000-0003-1292-1491}
\affiliation{
    \department{Information Retrieval and Knowledge Management Research Lab}
    \department{School of Information Technology}
    \institution{York University}
    \streetaddress{4700 Keele Street}
    \city{Toronto}
    \country{Canada}
    \postcode{M3J 1P3}}
\email{jhuang@yorku.ca}

\renewcommand{\shortauthors}{Lei Liu \& Jimmy Xiangji Huang}

\begin{abstract}
Dialogue systems for non-English languages have long been under-explored. In this paper, we take the first step to investigate few-shot cross-lingual transfer learning (FS-XLT) and multitask learning (MTL) in the context of open-domain dialogue generation for non-English languages with limited data. We observed \textit{catastrophic forgetting} in both FS-XLT and MTL for all 6 languages in our preliminary experiments. To mitigate the issue, we propose a simple yet effective prompt learning approach that can preserve the multilinguality of multilingual pre-trained language model (mPLM) in FS-XLT and MTL by bridging the gap between pre-training and fine-tuning with \textit{Fixed-prompt LM Tuning} and our hand-crafted prompts. Experimental results on all 6 languages in terms of both automatic and human evaluations demonstrate the effectiveness of our approach. Our code is available at \href{https://github.com/JeremyLeiLiu/XLinguDial}{https://github.com/JeremyLeiLiu/XLinguDial}.
\end{abstract}





\maketitle

\section{Introduction and motivation}
\label{sec:introduction}
Existing powerful dialogue models \cite{zhang-etal-2020-dialogpt,adiwardana-etal-2020-towards,roller-etal-2021-recipes,thoppilan-etal-2022-lamda} have been typically pre-trained on a significant number of English dialogue sessions extracted from either social media (e.g. Reddit and Twitter) or web documents. However, dialogue systems for many other languages have long been under-explored. An intuitive approach to building dialogue systems in other languages is to train from scratch on large-scale dialogue data of a specific language (e.g. Chinese \cite{bao-etal-2020-plato,bao-etal-2021-plato-2,bao-etal-2021-plato-xl,wang-etal-2022-a,zhou-etal-2021-eva,gu2023eva2} and Japanese \cite{sugiyama-etal-2021-empirical}), but it consumes considerable computing resources for each language, which is not eco-friendly. In addition, it is unlikely to collect massive dialogue sessions of some languages due to the data scarcity and/or scatteredness problem \cite{arora-etal-2022-computational}. Another line of investigation has focused on adaptation from pre-trained English models \cite{adewumi-etal-2021-smprat,adewumi-etal-2022-afriwoz}. Such adaptation, however, is far from satisfactory given significant vocabulary mismatch.

To address the problem of data unavailability, some prior work has focused on training a model on the dialogue sessions of another language translated from English \cite{naous-etal-2020-empathy}, but it involves much human efforts in post-editing. Others have utilized pre-trained language model (PLM) for the English language with the help of machine translation (MT), but MT pipeline is somewhat limited by translation quality, system turnaround, budget for APIs, and others \cite{lin-etal-2021-xpersona}. Another approach is to initialize a dialogue model with the learned parameters from off-the-shelf PLM for the same language (e.g. Arabic PLM \cite{antoun-etal-2020-arabert}) and then fine-tuning on a small-scale dialogue data \cite{naous-etal-2021-empathetic}. Unfortunately, PLM is not available for most languages due to the shortage of data and computing resources. Recent work has explored multilingual dialogue generation by fine-tuning mPLM \cite{conneau-etal-2020-unsupervised,chi-etal-2020-cross-lingual,xue-etal-2021-mt5,zhang-etal-2020-dialogpt} on dialogue data of multiple languages simultaneously, but the performance varies among different languages \cite{zhang-etal-2022-mdia}.

While mPLM has shown promise in enabling cross-lingual transfer for generation tasks, zero-shot cross-lingual transfer (ZS-XLT) with mPLM suffers much from \textit{catastrophic forgetting}, where mPLM that has been fine-tuned on the \textit{source language} is unable to generate fluent sentences in the \textit{target language} when being evaluated on it \cite{lauscher-etal-2020-zero,xue-etal-2021-mt5,maurya-etal-2021-zmbart,vu-etal-2022-overcoming}. In this work, we would like to investigate if \textit{catastrophic forgetting} occurs in two other cross-lingual scenarios namely FS-XLT and MTL in the context of dialogue generation given the availability of just a few data examples in \textit{target language}. We observed \textit{catastrophic forgetting} in our preliminary experiments of both scenarios. To mitigate the issue, we propose a prompt learning method that preserves the multilinguality of mPLM during fine-tuning with the help of hand-crafted prompts where we cast dialogue generation into the pre-training tasks of mPLM (e.g. \textit{span-corruption} in mT5 \cite{xue-etal-2021-mt5}). Experimental results on all 6 languages demonstrate our approach effectively mitigates the issue in both scenarios, thereby improving the overall generation quality.

Our main contributions are summarized in the following.

\begin{itemize}
    \item To the best of our knowledge, we are the first to investigate \textit{catastrophic forgetting} in the context of FS-XLT and MTL for open-domain dialogue generation, which sheds light on the study of non-English dialogue generation in the future.
    \item We propose a simple yet effective prompt learning method that preserves multilinguality of mPLM by bridging the gap between pre-training and fine-tuning to mitigate the issue, providing insights into cross-lingual transfer for generation.
    \item Experimentation with the \textit{Fixed-prompt LM Tuning} approach and our hand-crafted task-specific prompts on 6 languages.
\end{itemize}

\section{Methodology}
\subsection{Non-English Dialogue Generation}
\label{section-dialogue-generation}
Dialogue generation models are typically built on the sequence-to-sequence (seq2seq) model \cite{sutskever-etal-2014-sequence,DBLP:journals/corr/BahdanauCB14}, an encoder-decoder architecture where both the encoder and decoder can be either recurrent neural networks (RNNs) \cite{rumelhart-etal-1986-learning} or Transformers with self-attention blocks \cite{DBLP:conf/nips/VaswaniSPUJGKP17}. Let the input sequence be $X=(x_1, x_2,...,x_T)$ termed \textit{context} and the output sequence be $Y=(y_1,y_2,...,y_{T^{'}})$ termed \textit{response}, the learning objective of the task is to maximize the generation probability of \textit{response} conditioned on \textit{context}:
{\small
    \begin{equation}
    p(Y|X) = \prod_{t^{'}=1}^{T^{'}} p(y_{t^{'}}|y_1,y_2,...,y_{t^{'}-1},X)
    \end{equation}
}where $p(y_{t^{'}}|y_1,y_2,...,y_{t^{'}-1},X)$ denotes the conditional probability of $y_{t^{'}}$ given \textit{context} $X$ and its prior words in \textit{response} $Y$.

Existing work on English dialogue generation has mostly been done by training from scratch on massive English dialogues with the above objective \cite{zhang-etal-2020-dialogpt,adiwardana-etal-2020-towards,roller-etal-2021-recipes,thoppilan-etal-2022-lamda}, but that is not the case for most non-English languages due to a few factors introduced in prior sections. In this work, we approach to dialogue generation for non-English languages by exploring two cross-lingual transfer scenarios including FS-XLT and MTL with the help of mPLM (e.g. mT5 \cite{xue-etal-2021-mt5}) where we assume the availability of just a few data examples of those non-English languages.

\subsection{Few-shot Cross-lingual Transfer Learning}
\label{sec:few-shot-cross-lingual}
FS-XLT with the help of mPLM \cite{lauscher-etal-2020-zero,hedderich-etal-2020-transfer,zhao-etal-2021-closer} is a promising approach to many natural language processing (NLP) tasks for resource-lean languages where we assume the access to sufficient training data of \textit{source language} but just a few data examples (i.e. \textbf{few-shot}) of \textit{target language}. Generally, we have two stages in FS-XLT:

\begin{itemize}
    \item \textbf{Source-training}: mPLM is fine-tuned on the full training data of the \textit{source language} (e.g. English);
    \item \textbf{Target-adapting}: the source-trained model is further fine-tuned on the \textbf{few-shot} examples of the \textit{target language}.
\end{itemize}

\subsection{Multitask Learning}
\label{sec:multitask-learning}
MTL is an effective inductive transfer approach that improves generalization by jointly learning one or more \textit{auxiliary tasks} together with \textit{target task} \cite{DBLP:journals/ml/Caruana97,DBLP:journals/corr/LuongLSVK15,changpinyo-etal-2018-multi}. In this work, we focus on \textit{pairwise MTL} where there is only one \textit{auxiliary task} trained together with \textit{target task} as it works better when target dataset is smaller than auxiliary dataset \cite{weller-etal-2022-use}. When it comes to dialogue generation, we have \textbf{\textit{auxiliary language(s)}} and \textbf{\textit{target language}} as \textit{auxiliary task(s)} and \textit{target task}, respectively. Specifically, we have three stages here:

\begin{itemize}
    \item \textbf{Data interleaving}: we \textit{evenly} interleave the full training data of \textit{auxiliary language} with few-shot examples of \textit{target language}, with one data example of \textit{target language} inserted for every $N_{aux} / N_{tgt}$ data examples of \textit{auxiliary language} where $N_{aux}$ and $N_{tgt}$ denote the number of data examples in the training set of \textit{auxiliary language} and the number of \textbf{few-shot} data examples of the \textit{target language}, respectively;
    \item \textbf{Multitask training}: mPLM is fine-tuned on the interleaved data that contains \textit{context-response} pairs of both languages;
    \item \textbf{Target evaluation}: the fine-tuned mPLM is evaluated on the test set of \textit{target language}.
\end{itemize}

\subsection{Prompt Learning to Mitigate Catastrophic Forgetting in FS-XLT and MTL}
\label{sec:catastrophic-forgetting}
As is discussed in prior sections, zero-shot cross-lingual generation with mPLM \cite{lauscher-etal-2020-zero,maurya-etal-2021-zmbart} suffers much from \textit{catastrophic forgetting} \cite{xue-etal-2021-mt5,vu-etal-2022-overcoming}. Prior work assumes it is because mPLM has never been exposed to sentences of any other languages especially the \textit{target language} during its fine-tuning on supervised data of \textit{source language} \cite{xue-etal-2021-mt5}. To mitigate the issue, they mix a small amount of unsupervised data that covers either all the languages used in pre-training \cite{xue-etal-2021-mt5} or the \textit{target language} only \cite{vu-etal-2022-overcoming} into its fine-tuning on \textit{source language}.

Different from previous work, we explore whether \textit{catastrophic forgetting} occurs in two other cross-lingual transfer scenarios (i.e. FS-XLT and MTL) in dialogue generation task given the availability of just a few data examples of the \textit{target language}. In our preliminary experiments, we observed the issue in both FS-XLT and MTL, which is not surprising as we assume the availability of only 10 supervised data examples of each \textit{target language} in both \textbf{target-adapting} of FS-XLT and \textbf{multitask training} of MTL. Our intuition here is that the unsupervised pre-training tasks of multiple languages which essentially bring multilinguality to mPLM during the pre-training stage are not presented in the fine-tuning stage (i.e. FS-XLT or MTL), making it difficult for mPLM to maintain their multilinguality.

Inspired by the prompt learning approaches with PLM for the English language \cite{liu-etal-2023-pre-train,gao-etal-2021-making,schick-schutze-2021-just,schick-schutze-2021-exploiting,yuan2021bartscore,DBLP:journals/corr/abs-2110-08118,kasahara-etal-2022-building}, we propose to preserve the multilinguality of mPLM by bridging the gap between the unsupervised pre-training tasks and downstream tasks (e.g. dialogue generation) in the fine-tuning stage (i.e. FS-XLT and MTL). Specifically, we cast dialogue generation into the pre-training tasks of mPLM to bridge the gap, thereby maintaining the multilinguality of mPLM during fine-tuning and mitigating \textit{catastrophic forgetting}. While there are a few mPLMs available in the literature \cite{xue-etal-2021-mt5,liu-etal-2020-multilingual-denoising,chi-etal-2021-mt6}, in this work we experiment with mT5 \cite{xue-etal-2021-mt5} only as we aim to shed light on the study of non-English dialogue generation, instead of establishing strong baselines or state-of-the-art performances.

Considering the \textbf{span corruption} task in the pre-training stage of mT5, we use the first two sentinel tokens namely \textbf{<extra\_id\_0>} and \textbf{<extra\_id\_1>} to perform \textbf{span corruption} on all the \textit{context-response} pairs where our hand-crafted task prompts "\textbf{Context:}" and "\textbf{Response:}" are utilized to format each \textit{context-response} pair. Let \textit{context} be $[X]$, \textit{response} be $[Y]$, we formulate dialogue generation in Table \ref{tab:manual-prompts}. In our experiments, we adopt the \textit{Fixed-prompt LM Tuning} approach \cite{schick-schutze-2021-exploiting,schick-schutze-2021-shot,gao-etal-2021-making} with our task-specific prompts where we do not introduce additional parameters for prompts and only update the parameters of mPLM during fine-tuning (i.e. FS-XLT and MTL).

\begin{table}[t]
    \small
    \caption{Hand-crafted task-specific prompts.}
    \label{tab:manual-prompts}
    \begin{tabular}{c c}
        \toprule
        
        \textit{Context}/\textit{Response} & Data Format \\ 
        \midrule
        
        \textit{Context} & \textbf{Context:}\ [X]\ \textbf{Response:} \ \textbf{<extra\_id\_0>} \\
        \textit{Response} & \textbf{<extra\_id\_0>}\ [Y]\ \textbf{<extra\_id\_1>} \\
        
        \bottomrule
    \end{tabular}
\end{table}

\section{Experiments}

\subsection{Dataset}
To the best of our knowledge, MDIA \cite{zhang-etal-2022-mdia} is the only publicly available multilingual benchmark for the dialogue generation task, so we simply adopt it for experimentation. While MDIA covers more than 40 languages, in this work, we focus on a few select languages for our preliminary investigation due to space limitations. As a common practice, \textbf{English} is taken as the \textit{source}/\textit{auxiliary language} in FS-XLT/MTL. We consider Danish (da), German (de) and Norwegian (no) as the representatives of \textbf{Germanic} language genus along with Spanish (es), Italian (it) and Portuguese (pt) as the representatives of \textbf{Romance} language genus\footnote{The Germanic/Romance language genus is termed in the \href{https://wals.info/}{World Atlas of Language Structures (WALS)} database.}, and adopt each one of them as the \textit{target language} in both FS-XLT and MTL.

The data statistics of all 6 non-English languages together with English are shown in Table \ref{tab:data-statistics} where the values of last three columns denote the number of \textit{context-response} pairs in the training (train), validation (valid) and test set, respectively. Specifically, data examples in the training set for each non-English language are randomly chosen from their corresponding training set in MDIA, and are taken as the \textbf{few-shot} data in the \textbf{target-adapting} stage of FS-XLT. Note that a fixed random seed is used when choosing those 10 examples to ensure the reproducibility of results presented in this work. On top of it, the full training data of English is evenly interleaved with those 10 data examples for each individual non-English language respectively, with one data example of the non-English language inserted for every 1000 English data examples. The interleaved data is used in the \textbf{multitask training} stage of MTL.

\begin{table}[t]
    \caption{Statistics of all 6 languages used in this work.}
    \footnotesize
    \begin{tabular}{c c c c c c}
        \toprule
        \textbf{ISO-639-1} & \textbf{Language} & \textbf{Genus} & \textbf{\# Train} & \textbf{\# Valid} & \textbf{\# Test} \\ 
        \midrule
        en & English & Germanic & 10000 & 1000 & 1000 \\ \cdashline{1-6}
        da & Danish & Germanic & 10 & 1000 & 1000 \\
        de & German & Germanic & 10 & 1000 & 1000 \\
        no & Norwegian & Germanic & 10 & 1000 & 1000 \\
        es & Spanish & Romance & 10 & 1000 & 1000 \\
        it & Italian & Romance & 10 & 1000 & 1000 \\
        pt & Portuguese & Romance & 10 & 1000 & 1000 \\
        \bottomrule
    \end{tabular}
    \label{tab:data-statistics}
\end{table}

\begin{table}[t]
    \caption{Automatic evaluation on all languages.}
    \footnotesize
    \begin{tabular}{c c c c c c c c}
        \toprule
        
        \multirow{2}{*}{\textbf{Lang}} & \multirow{2}{*}{\textbf{Setting}} & \multicolumn{4}{c}{\textbf{sacreBLEU}} & \multicolumn{2}{c}{\textbf{Distinct-N}} \\
        & & \textbf{B-1} & \textbf{B-2} & \textbf{score} & \textbf{bp} & \textbf{D-1} & \textbf{D-2} \\
        \midrule
        
        en & FT & 15.77 & 1.08 & 0.31 & 0.39 & 9.52 & 24.07 \\ \hline
        
        \multirow{4}{*}{da} & FS-XLT & 11.29 & 0.68 & 0.20 & 0.73 & 6.35 & 17.34 \\ & FS-XLT$_{pmpt}$ & \textbf{12.51} & \textbf{0.77} & \textbf{0.23} & 0.47 & \textbf{9.80} & \textbf{27.92} \\ \cdashline{2-8}
        & MTL & 13.28 & 0.83 & 0.28 & 0.53 & 7.48 & 22.03 \\ & MTL$_{pmpt}$ & \textbf{14.76} & \textbf{1.17} & \textbf{0.33} & 0.35 & \textbf{10.75} & \textbf{32.17} \\ \hline
        
        \multirow{4}{*}{de} & FS-XLT & 4.63 & 0.21 & 0.11 & 0.85 & 7.93 & 20.16 \\ & FS-XLT$_{pmpt}$ & \textbf{4.84} & \textbf{0.36} & \textbf{0.12} & 0.66 & \textbf{12.68} & \textbf{32.40} \\ \cdashline{2-8}
        & MTL & 5.34 & 0.12 & 0.05 & 1.00 & 7.14 & 18.00 \\ & MTL$_{pmpt}$ & \textbf{8.85} & \textbf{0.51} & \textbf{0.29} & 1.00 & \textbf{10.83} & \textbf{30.06} \\ \hline
        
        \multirow{4}{*}{no} & FS-XLT & 12.27 & 0.67 & 0.19 & 0.66 & 4.70 & 13.22 \\ & FS-XLT$_{pmpt}$ & \textbf{13.54} & \textbf{1.11} & \textbf{0.25} & 0.59 & \textbf{6.09} & \textbf{17.02} \\ \cdashline{2-8}
        & MTL & 14.45 & 0.81 & 0.13 & 0.43 & 7.18 & 19.52 \\ & MTL$_{pmpt}$ & \textbf{17.74} & \textbf{1.70} & \textbf{0.37} & 0.31 & \textbf{10.39} & \textbf{29.13} \\ \hline
        
        \multirow{4}{*}{es} & FS-XLT & 6.24 & 0.31 & 0.24 & 1.00 & 4.54 & 11.74 \\ & FS-XLT$_{pmpt}$ & \textbf{8.21} & \textbf{0.52} & \textbf{0.32} & 0.82 & \textbf{6.50} & \textbf{16.43} \\ \cdashline{2-8}
        & MTL & 4.88 & 0.25 & 0.18 & 0.83 & 3.12 & 7.14 \\ & MTL$_{pmpt}$ & \textbf{9.37} & \textbf{0.68} & \textbf{0.30} & 0.62 & \textbf{8.48} & \textbf{22.14} \\ \hline
        
        \multirow{4}{*}{it} & FS-XLT & 3.63 & 0.23 & 0.18 & 1.00 & 5.77 & 13.83 \\ & FS-XLT$_{pmpt}$ & \textbf{5.12} & \textbf{0.53} & \textbf{0.24} & 0.76 & \textbf{9.24} & \textbf{20.81} \\ \cdashline{2-8}
        & MTL & 4.72 & 0.34 & 0.09 & 0.50 & 8.51 & 19.44 \\ & MTL$_{pmpt}$ & \textbf{7.36} & \textbf{0.75} & \textbf{0.47} & 0.63 & \textbf{9.52} & \textbf{23.99} \\ \hline
        
        \multirow{4}{*}{pt} & FS-XLT & 7.34 & 0.35 & 0.15 & 1.00 & 6.18 & 15.98 \\ & FS-XLT$_{pmpt}$ & \textbf{8.85} & \textbf{0.79} & \textbf{0.47} & 0.92 & \textbf{8.65} & \textbf{22.30} \\ \cdashline{2-8}
        & MTL & 7.72 & 0.47 & 0.10 & 0.45 & 6.69 & 15.41 \\ & MTL$_{pmpt}$ & \textbf{12.97} & \textbf{0.76} & \textbf{0.15} & 0.60 & \textbf{8.78} & \textbf{21.07} \\
        
        \bottomrule
    \end{tabular}
    \label{tab:automatic-evaluation}
\end{table}

\subsection{Evaluation Metrics}
\noindent \textbf{Automatic evaluation.} We use SacreBLEU\footnote{\href{https://github.com/mjpost/sacreBLEU}{https://github.com/mjpost/sacreBLEU}} \cite{post-2018-call} to measure the word overlap between predicted responses and reference responses as it produces shareable, comparable, and reproducible BLEU \cite{papineni-etal-2002-bleu} scores. In addition, we use Distinct N-grams \cite{li-etal-2016-diversity} to measure the diversity of predicted responses. \textbf{Human Evaluation.} We conduct human evaluations on Prolific\footnote{\href{https://www.prolific.co/}{https://www.prolific.co/}} to further verify the effectiveness of our proposed method. We recruit one annotator for each language given the limited number of native speakers of some languages available. Systems are paired with each other based on whether they are under the same setting (e.g. FS-XLT is paired with FS-XLT$_{pmpt}$). We randomly pick 100 samples from the test set of each language. For each sample, given the \textit{context}, a native speaker of the corresponding language is asked to indicate the \textit{response} they believe is better than the other in terms of \textit{Fluency} which pertains to the overall grammatical correctness and fluidity of a \textit{response}. In cases where it is difficult to determine which \textit{response} is better, annotators can indicate “Neutral”.

\begin{table}[t]
    \caption{Human evaluations in terms of \textit{Fluency}.}
    \footnotesize
    \begin{tabular}{c c c c c c c}
        \toprule
    
        \textbf{Lang} & \vline & \textbf{Setting} & \textbf{Fluency} & \vline & \textbf{Setting} & \textbf{Fluency} \\
        \midrule
        
        \multirow{3}{*}{da} & \vline & FS-XLT & 29 (29\%) & \vline & MTL & 30 (30\%) \\ & \vline & Neutral & 9 (9\%) & \vline & Neutral & 10 (10\%) \\ & \vline & FS-XLT$_{pmpt}$ & \textbf{62 (62\%)} & \vline & MTL$_{pmpt}$ & \textbf{60 (60\%)} \\
        \midrule
        
        \multirow{3}{*}{de} & \vline & FS-XLT & 40 (40\%) & \vline & MTL & 18 (18\%) \\ & \vline & Neutral & 9 (9\%) & \vline & Neutral & 27 (27\%) \\ & \vline & FS-XLT$_{pmpt}$ & \textbf{51 (51\%)} & \vline & MTL$_{pmpt}$ & \textbf{55 (55\%)} \\
        \midrule
        
        \multirow{3}{*}{no} & \vline & FS-XLT & \textbf{49 (49\%)} & \vline & MTL & 25 (25\%) \\ & \vline & Neutral & 2 (2\%) & \vline & Neutral & 8 (8\%) \\ & \vline & FS-XLT$_{pmpt}$ & \textbf{49 (49\%)} & \vline & MTL$_{pmpt}$ & \textbf{67 (67\%)} \\
        \midrule
        
        \multirow{3}{*}{es} & \vline & FS-XLT & 23 (23\%) & \vline & MTL & 17 (17\%) \\ & \vline & Neutral & 36 (36\%) & \vline & Neutral & 33 (33\%) \\ & \vline & FS-XLT$_{pmpt}$ & \textbf{41 (41\%)} & \vline & MTL$_{pmpt}$ & \textbf{50 (50\%)} \\
        \midrule
        
        \multirow{3}{*}{it} & \vline & FS-XLT & 23 (23\%) & \vline & MTL & 22 (22\%) \\ & \vline & Neutral & 37 (37\%) & \vline & Neutral & 23 (23\%) \\ & \vline & FS-XLT$_{pmpt}$ & \textbf{40 (40\%)} & \vline & MTL$_{pmpt}$ & \textbf{55 (55\%)} \\
        \midrule
        
        \multirow{3}{*}{pt} & \vline & FS-XLT & 30 (30\%) & \vline & MTL & 2 (2\%) \\ & \vline & Neutral & 27 (27\%) & \vline & Neutral & 13 (13\%) \\ & \vline & FS-XLT$_{pmpt}$ & \textbf{43 (43\%)} & \vline & MTL$_{pmpt}$ & \textbf{85 (85\%)} \\
        
        \bottomrule
    \label{tab:human-evaluation}
    \end{tabular}
\end{table}

\begin{table*}[t]
    \caption{Examples of responses generated in different settings. Illegal responses are red-colored.}
    \footnotesize
    \begin{tabular}{c c c c c c}
        \toprule
        
        \textbf{Lang} & \textbf{Context} & \textbf{Setting} & \textbf{Response} & \textbf{Explanation} \\
        \midrule
        
        \multirow{4}{*}{da} & \multirow{4}{*}{Er dum eller døv} 
        & FS-XLT & {\color{red} I wouldn't argue anyone would have a girlfriend.} & {\color{red}\ding{55}} {\verb|Translated to English & Ungrammatical.|} \\
        & & FS-XLT$_{pmpt}$ & Ikke dum, men døv. & \ding{51} {\verb|Fluent in Danish.|} \\ \cline{3-5}
        & & MTL & {\color{red} I wouldn't say that.} & {\color{red}\ding{55}} {\verb|Translated to English.|} \\ 
        & & MTL$_{pmpt}$ & Så hvor er du dum? & \ding{51} {\verb|Grammatically correct in Danish.|} \\
        \midrule
    
        \multirow{4}{*}{de} & \multirow{4}{*}{Wann bist du eingestiegen?} 
        & FS-XLT & {\color{red} I was a few weeks ago a few weeks ago.} & {\color{red}\ding{55}} {\verb|Translated to English.|} \\ 
        & & FS-XLT$_{pmpt}$ & Also wann? & \ding{51} {\verb|Fluent in German.|} \\ \cline{3-5} 
        & & MTL & {\color{red} I'm sorry, I'm sorry.} & {\color{red}\ding{55}}  {\verb|Translated to English.|} \\
        & & MTL$_{pmpt}$ & Also, wenn du geht, geht du auch runter. & \ding{51} {\verb|All words are in German.|} \\
        \midrule
        
        \multirow{4}{*}{no} & \multirow{4}{*}{\makecell[lt]{Som oftest, men ikke \\ alltid. Noen yrker har \\ lovpålagt minstelønn.}}
        & FS-XLT & {\color{red}I skulle ønske at dette blir i løpet av året.} & {\color{red}\ding{55}} {\verb|Ungrammatical.|} \\
        & & FS-XLT$_{pmpt}$ & Ikke alle yrker har minstelønn. & \ding{51} {\verb|Fluent in Norwegian.|} \\ \cline{3-5}
        & & MTL & {\color{red}I skjønner det i skjønn.} & {\color{red}\ding{55}} {\verb|Ungrammatical.|}\\
        & & MTL$_{pmpt}$ & Ikke alle yrker har minstelønn. & \ding{51} {\verb|Fluent in Norwegian.|} \\
        
        \bottomrule
    \end{tabular}
    \label{tab:examples-of-responses}
\end{table*}

\subsection{Implementation Details}
In our experimentation, all models are implemented based on the mT5-base model from HuggingFace Transformers\footnote{\href{https://github.com/huggingface/transformers/tree/v4.8.0}{https://github.com/huggingface/transformers/tree/v4.8.0}} \cite{wolf-etal-2020-transformers}. The default tokenizer for mT5-base is used for tokenization. The \textit{contexts} and \textit{responses} whose lengths are longer than 64 are truncated. AdamW \cite{loshchilov2018decoupled} is used for optimization in model training with \begin{math} \beta_1 = 0.9, \beta_2 =0.999 \end{math} and \begin{math} \epsilon = 1e-8 \end{math}. For \textbf{source-training} in FS-XLT and \textbf{multitask training} in MTL, the mT5-base model is trained on the full training data of English or the interleaved data for up to 5 epochs with a batch size of 8 and the initial learning rate $lr=5e-5$, respectively. When it comes to the \textbf{target-adapting} in FS-XLT, source-trained model is fine-tuned on the \textbf{few-shot} data of \textit{target language} for 6 epochs with the batch size being 4 and initial learning rate $lr=1e-4$. The checkpoint (i.e. training step) with the minimum validation loss is selected as the best model in all the above settings. It is also worth noting that the validation set of \textit{auxiliary language} is used for model selection in the \textbf{multitask training} stage of MTL. All the experiments are conducted on 4 NVIDIA V100 GPUs.

\section{Results and Analysis}
We present automatic evaluation of both cross-lingual scenarios on all 6 non-English languages along with English fine-tuning (i.e. FT) in Table \ref{tab:automatic-evaluation}. Human evaluation are presented in Table \ref{tab:human-evaluation}. FS-XLT$_{pmpt}$ and MTL$_{pmpt}$ denote applying our prompt learning approach in Section \ref{sec:catastrophic-forgetting} to FS-XLT and MTL. We provide examples of responses generated in all the settings for 3 languages in Table \ref{tab:examples-of-responses}.

\subsection{Catastrophic Forgetting in FS-XLT and MTL}
As we can see in Table \ref{tab:automatic-evaluation}, the performances of FS-XLT and MTL on all the \textit{target languages} in terms of both sacreBLEU and Distinct-N are much inferior to that of English FT, which is not surprising, given that only 10 \textit{context-response} pairs of each \textit{target language} are available. As is illustrated in Table \ref{tab:examples-of-responses}, \textit{catastrophic forgetting} is observed on each \textit{target language} in FS-XLT and MTL where the model tends to generate illegal responses that are either in English in their entirety or ungrammatical in the \textit{target language}.

\subsection{Prompt Learning Improves FS-XLT and MTL}
As we can see from Table \ref{tab:automatic-evaluation} and \ref{tab:human-evaluation}, prompt learning approaches including FS-XLT$_{pmpt}$ and MTL$_{pmpt}$ outperform their counterparts (i.e. FS-XLT and MTL) in terms of both automatic and human evaluation across all \textit{target languages}, demonstrating the effectiveness of our \textit{Fixed-prompt LM Tuning} approach with hand-crafted task-specific prompts in preserving the multilinguality of mPLM during fine-tuning and mitigating \textit{catastrophic forgetting}. It is worth noting that the reason why the performance of FS-XLT$_{pmpt}$ is comparable to (instead of better than) FS-XLT for Norwegian (no) was that the annotator did not strictly follow our instructions in the beginning and thus made a number of selections which are not justified.

Most significantly, we observe \textit{catastrophic forgetting} has been greatly mitigated in both FS-XLT$_{pmpt}$ and MTL$_{pmpt}$. As is shown in Table \ref{tab:examples-of-responses}, compared to FS-XLT and MTL that wrongly generate English responses in their entirety (i.e. Danish and German) or ungrammatical responses (i.e. Norwegian), prompt learning approaches (i.e. FS-XLT$_{pmpt}$ and MTL$_{pmpt}$) tend to generate responses which are more fluent and grammatically correct in \textit{target language}. While we cannot provide examples for other languages (Spanish, Italian and Portuguese) due to space limitations, similar conclusions have been drawn for these languages. Overall, the results suggest our proposed approach mitigates \textit{catastrophic forgetting} and generates responses that are more fluent and grammatically correct.

\section{Conclusions and Future Work}
In this work, we investigated the problem of \textit{catastrophic forgetting} occurred in FS-XLT and MTL in the context of dialogue generation for non-English languages. We proposed a prompt learning method that preserves the multilinguality of mPLM during fine-tuning to mitigate the issue. Results on all 6 languages demonstrate the effectiveness of our proposed approach. We will explore whether the proposed approach mitigates \textit{catastrophic forgetting} in FS-XLT and MTL for other languages in our future work. In addition, we would like to investigate how much the model performance varies when using different task prompts in the future. It is also interesting to explore soft/trainable prompts for dialogue generation.

\begin{acks}
This research is supported by the Natural Sciences and Engineering Research Council (NSERC) of Canada, the York Research Chairs (YRC) program and an ORF-RE (Ontario Research Fund Research Excellence) award in BRAIN Alliance. Computations were made on the supercomputer Béluga, managed by Calcul Québec and the Digital Research Alliance of Canada.
\end{acks}

\bibliographystyle{ACM-Reference-Format}
\balance
\bibliography{sigir23-leiliu}










\end{document}